\documentclass[twoside,leqno,twocolumn]{article}

\usepackage{dblfloatfix}

\usepackage[letterpaper]{geometry}

\usepackage{siamproceedings}

\makeatletter
\AddToHook{env/algorithm/begin}{%
  \if@twocolumn
    \setlength{\textwidth}{\columnwidth}%
  \fi
}%
\makeatother

\usepackage[T1]{fontenc}
\usepackage{amsfonts}
\usepackage{graphicx}
\usepackage{epstopdf}
\usepackage{enumitem}
\usepackage{algorithmic}
\ifpdf
  \DeclareGraphicsExtensions{.eps,.pdf,.png,.jpg}
\else
  \DeclareGraphicsExtensions{.eps}
\fi

\usepackage{soul}
\usepackage{url}
\usepackage{float}
\hypersetup{hidelinks}
\usepackage{placeins}
\usepackage[protrusion=true,expansion=false]{microtype}
\usepackage[utf8]{inputenc}
\usepackage[small]{caption}
\usepackage{amsmath}
\usepackage{nccmath}
\usepackage{amssymb}
\usepackage{booktabs}
\usepackage{dsfont}
\usepackage{tcolorbox}
\usepackage{xspace}
\usepackage{multirow}

\newcommand\relatedversion{}
\renewcommand\relatedversion{\thanks{The full version of the paper can be accessed at \protect\url{https://arxiv.org/abs/2602.17686}}}

\title{\Large BRIDGE: Bridging Reasoning In Distillation Gap Elimination\\ via Structure-Aware Masking\relatedversion}

\author{Bowen Yu\thanks{City University of Hong Kong (\email{bowyu2-c@my.cityu.edu.hk}, \email{szhang844-c@my.cityu.edu.hk}, \email{binhawang2-c@my.cityu.edu.hk}, \email{wenyiwy2022@my.cityu.edu.hk}, \email{jt.g@my.cityu.edu.hk}, \email{boweliu6-c@my.cityu.edu.hk}, \email{zmzhao6-c@my.cityu.edu.hk}, \email{wanyuwang4-c@my.cityu.edu.hk}).}
\and Sheng Zhang\footnotemark[2]
\and Binhao Wang\footnotemark[2]
\and Yi Wen\footnotemark[2]
\and Jingtong Gao\footnotemark[2]
\and Bowen Liu\footnotemark[2]
\and Zimo Zhao\footnotemark[2]
\and Shanshan Ye\thanks{Mohamed Bin Zayed University of Artificial Intelligence (\email{shanshan.ye@mbzuai.ac.ae}).}
\and Wanyu Wang\footnotemark[2]
\and Maolin Wang\thanks{Corresponding authors: City University of Hong Kong (\email{Morin.wang@my.cityu.edu.hk}, \email{xianzhao@cityu.edu.hk}).}
\and Xiangyu Zhao\footnotemark[4]}

\date{}

\begin{document}

\maketitle

\fancyfoot[R]{\scriptsize{Copyright \textcopyright\ 2026 by SIAM\\
Unauthorized reproduction of this article is prohibited}}

\begin{abstract}
Chain-of-Thought (CoT) reasoning has significantly improved LLMs' mathematical problem-solving capabilities, but distilling such capabilities into smaller models remains challenging due to the capacity mismatch between verbose teachers and compact students. Directly copying teachers' lengthy reasoning chains causes capacity overload, resulting in truncated outputs or repetitive failure. Existing remedies each sacrifice a critical property of CoT: implicit reasoning methods (e.g., compressing reasoning into hidden states) trade away interpretability and verifiability, while heuristic compression strategies (e.g., random step pruning) destroy logical integrity. To address this, we propose BRIDGE, a curriculum framework that first establishes structural understanding via masked reconstruction, then uses GRPO-based reinforcement learning to guide students in self-discovering the optimal balance between accuracy and brevity, and finally internalizes complex reasoning through teacher-guided rewriting on failure cases. On GSM8K, BRIDGE enables Qwen2.5-3B to achieve 11.29\% accuracy improvement and 27.4\% token reduction over the original model, outperforming instruction-tuned variants and distillation baselines. Zero-shot transfer experiments on SVAMP and MATH-500 further confirm the generalization of internalized reasoning. Our code and model checkpoints are publicly available at \url{https://github.com/Applied-Machine-Learning-Lab/SDM2026_BRIDGE} and \url{https://huggingface.co/bowen0815/BRIDGE}.
\end{abstract}

\section{Introduction.}

\begin{figure*}[!t]
    \centering
    \includegraphics[width=0.95\textwidth]{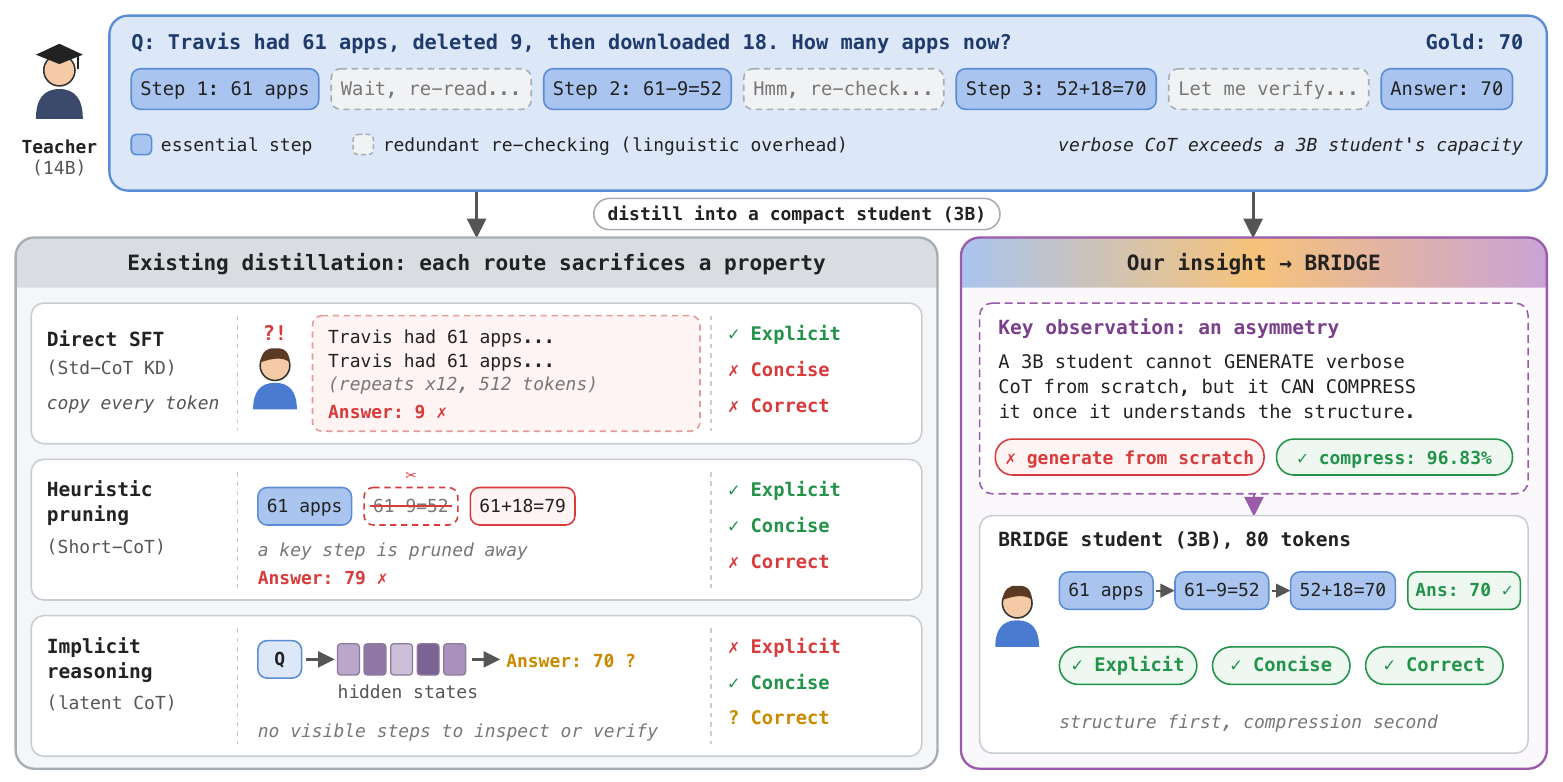}
    \caption{Motivation of \textbf{BRIDGE}. When distilling a verbose teacher CoT into a 3B student, each existing route sacrifices one property of CoT: direct SFT overloads the student into repetition, heuristic pruning drops key steps, and implicit reasoning hides the process. \textbf{BRIDGE} builds on the asymmetry that students can compress reasoning they cannot generate, yielding explicit, concise and correct CoT. The pruning example is illustrative.}
    \label{fig:motivation}
\end{figure*}

Chain-of-Thought (CoT) reasoning has emerged as a transformative technique for eliciting complex problem-solving capabilities in large language models (LLMs)~\cite{zhang2026memsearch,zhang2026r2searchercalibratingretrievalreasoning,wang2026function,wang2026renormalization}. By prompting models to decompose tasks into explicit intermediate steps (e.g., via Chain-of-Thought exemplars), CoT enables models to tackle arithmetic and symbolic reasoning challenges with remarkable success~\cite{gao2025navigate,jia2025bridging,peng2025stepwise,ni2025zeroed,xu2025align,zhang2026process,zhang2025deep,liu2023exploration}. For instance, prompting strategies have been shown to boost GSM8K accuracy from 17.9\% to 58.1\% in few-shot settings~\cite{wei2022chain} and from 10.4\% to 40.7\% in zero-shot scenarios~\cite{kojima2022large}. However, these gains are predominantly observed in massive models (e.g., models with tens of billions of parameters). Deploying such capabilities in resource-constrained environments requires distilling them into smaller models (e.g., 3B parameters), a process that remains computationally challenging.

A fundamental obstacle in this distillation process is the \textbf{capacity mismatch} between teacher and student~\cite{li2025small}. Capable teachers (e.g., DeepSeek-R1-14B) often rely on lengthy reasoning chains to ensure correctness. When compact students (e.g., 3B models) attempt to reproduce these lengthy sequences via standard supervised fine-tuning, they lack the representational bandwidth to process or memorize such content effectively. This manifests as truncated outputs, repetition loops, or superficial mimicry without genuine understanding~\cite{magister2023teaching}. For example, a student trained by standard SFT repeats the question twelve times and outputs a wrong answer (Figure~\ref{fig:motivation}, left).
 
Several approaches have been proposed to address this mismatch, yet none satisfies the need for explicit, verifiable reasoning. Implicit reasoning methods~\cite{deng2023implicit,li2025implicit,shen2025codi} bypass sequence length constraints by compressing reasoning into hidden states or continuous representations. However, this sacrifices interpretability and verifiability, the very properties that make CoT valuable for debugging and auditing. Heuristic compression strategies~\cite{xia2025tokenskip,li2023mixed} attempt to shorten rationales via random pruning or mixed-length training, but such aggressive truncation destroys logical integrity, resulting in incomplete reasoning chains with degraded readability. As illustrated in Figure~\ref{fig:motivation}, each existing route sacrifices one critical property of CoT: direct imitation overloads the student into repetition, heuristic pruning breaks the logical chain, and implicit reasoning leaves nothing to verify. A \textit{critical challenge} remains: How can we enable a small model to maintain explicit, verifiable reasoning while compressing it to fit within its limited capacity? This fundamentally requires mining effective reasoning substructures from verbose teacher CoT: identifying the minimal yet sufficient reasoning steps that a compact student model can genuinely internalize.

Our key observation is an asymmetry: although a compact student cannot generate verbose CoT from scratch, it can compress the teacher's reasoning once given its structure. In a preliminary study, 96.83\% of the student's failure cases admit a valid compression when the teacher's CoT is provided (Figure~\ref{fig:motivation}, right). Building on this observation, we propose \textbf{BRIDGE}, a curriculum learning framework built on the premise that \textbf{effective compression requires structural understanding}. Rather than forcing students to memorize verbose chains immediately, \textbf{BRIDGE} first establishes a structural foundation through masked reconstruction, training the student to recognize logical dependencies. It then employs Group Relative Policy Optimization (GRPO) on masked completion tasks to guide the student in self-discovering the optimal balance between accuracy and brevity. In the final stage, for difficult queries where the student struggles, it utilizes teacher-guided rewriting to progressively and effectively internalize complex reasoning deeply into concise form.
We summarize our main contributions as follows:

(1) We identify capacity mismatch as the primary bottleneck in reasoning distillation and show that direct SFT on verbose CoT is detrimental to small models.

(2) We introduce \textbf{BRIDGE}, a structure-aware curriculum framework that enables compact models to internalize and rewrite reasoning chains efficiently.

(3) We demonstrate that Qwen2.5-3B trained with BRIDGE achieves 11.29\% accuracy improvement and 27.4\% output length reduction on GSM8K, outperforming instruction-tuned variants and distillation baselines.

\section{Methodology.}

\begin{figure*}[!t]
    \centering
    \includegraphics[width=0.9\textwidth]{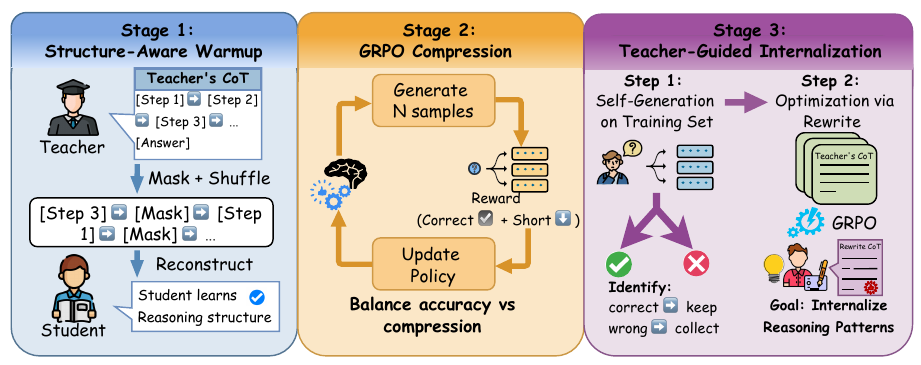}
    \caption{Overview of \textbf{BRIDGE}. Stage 1 establishes structural understanding through masked shuffled reconstruction. Stage 2 applies GRPO on masked completion tasks to balance accuracy and compression. Stage 3 identifies failure cases, applies teacher-guided rewriting for internalization, and uses GRPO to maintain compression capabilities.}
    \label{fig:framework}
\end{figure*}

We propose \textbf{BRIDGE}, a three-stage curriculum learning framework that progressively builds the student's reasoning capabilities. Figure~\ref{fig:framework} illustrates the overall architecture.

\subsection{Framework Overview.}

Given a question $q$ sampled from dataset $\mathcal{D}$, a teacher model with policy $\pi_{\mathcal{T}}$ generates a verbose chain-of-thought response $r_{\mathcal{T}} = (s_1, s_2, \ldots, s_n)$ consisting of $n$ reasoning steps. Our goal is to train a compact student policy $\pi_\theta$ that produces correct but concise reasoning.

The design follows a curriculum learning principle: rather than directly optimizing for brevity (which often leads to reward hacking on weak models~\cite{ouyang2022training}), we first establish structural competence (Stage 1), then introduce length constraints via exploration (Stage 2), and finally internalize knowledge from hard cases while preserving compression incentives (Stage 3).

We formalize the \textbf{BRIDGE} training procedure in \textbf{Algorithm~\ref{alg:bridge}}. The curriculum progressively builds the student's capabilities: \textbf{Stage 1} teaches structural reasoning via reconstruction, \textbf{Stage 2} optimizes the accuracy-compression trade-off via GRPO on masked completion tasks, and \textbf{Stage 3} internalizes teacher knowledge for difficult samples while maintaining brevity incentives.

\begin{algorithm}[H]
    \caption{\textbf{BRIDGE} Training Curriculum}
    \label{alg:bridge}
    \footnotesize
    \begin{algorithmic}[1]
    \setlength{\itemsep}{1pt}
    \REQUIRE Teacher $\pi_{\mathcal{T}}$, Student $\pi_\theta$, Dataset $\mathcal{D}$
    \STATE \textbf{Stage 1: Structure-Aware Warmup}
    \FOR{batch $(q, r_{\mathcal{T}}) \sim \mathcal{D}$}
        \STATE Construct $\tilde{r}$ via Step Shuffling (100\%) + Stochastic Step Masking ($p=0.7$)
        \STATE Update $\pi_\theta$ via reconstruction loss $\mathcal{L}_{\text{Stage1}}$
    \ENDFOR
    \STATE \textbf{Stage 2: GRPO Compression}
    \WHILE{performance improving}
        \STATE Construct $\tilde{r}_{\text{S2}}$ via Step Masking (No Shuffling)
        \STATE Sample $G$ outputs $\{r_i\} \sim \pi_\theta(\cdot|\tilde{r}_{\text{S2}})$
        \STATE Compute Hierarchical Reward $R(r_i)$
        \STATE Update $\pi_\theta$ via GRPO with KL (Ref: $\pi_{\text{S1}}$)
    \ENDWHILE
    \STATE \textbf{Stage 3: Teacher-Guided Internalization}
    \STATE Identify failure cases $\mathcal{D}_{\text{hard}}$ from Stage 2
    \STATE Update $\pi_\theta$ to compress $r_{\mathcal{T}}$ via GRPO with KL (input: $q + r_{\mathcal{T}}$, Ref: $\pi_{\text{S2}}$)
    \RETURN Optimized $\pi_\theta$
    \end{algorithmic}
    \end{algorithm}

\subsection{Stage 1: Structure-Aware Capacity Warmup.}\label{sec:stage1}

The standard distillation objective trains the student to minimize the negative log-likelihood of reproducing the teacher's response:
\begin{equation}
    \mathcal{L}_{\text{SFT}} = -\mathbb{E}_{(q, r_{\mathcal{T}}) \sim \mathcal{D}} \left[ \log \pi_\theta(r_{\mathcal{T}} \mid q) \right].
\end{equation}
When $r_{\mathcal{T}}$ is lengthy, this approach often fails: the student lacks sufficient capacity to memorize extended sequences, leading to truncated outputs, repetitive patterns, or superficial mimicry. A primary cause is the student model's inability to capture long-term dependencies in detailed teacher outputs. Direct SFT forces the student to attend to local token patterns, often missing the global reasoning structure.

To address this, we propose a pre-training objective that builds a ``logical skeleton'' before refining details. Inspired by denoising objectives~\cite{vincent2008extracting,lewis2020bart} but adapted for reasoning structures, we introduce a \textbf{Structure-Aware Reconstruction} task. Given a teacher response $r_{\mathcal{T}} = (s_1, s_2, \ldots, s_n)$, we apply two transformations:

\paragraph{Step Shuffling.} The core insight is that sequential copying allows students to exploit positional shortcuts without understanding logic. To eliminate this shortcut, we apply shuffling to \textbf{all samples}, permuting the order of steps:
\begin{equation}
    (s_1, \ldots, s_n) \rightarrow (s_{\pi(1)}, \ldots, s_{\pi(n)}), \quad \pi \sim \text{Perm}(n).
\end{equation}
Restoring the correct order requires the student to recognize causal dependencies between steps, forcing it to understand the global semantic structure of the reasoning chain~\cite{he2022masked} rather than relying on local context alone.

\paragraph{Step Masking.} While shuffling tests ordering ability, a student might still restore order by matching surface-level keywords without truly understanding the logical connections. To address this, we additionally mask approximately 15\% of the reasoning steps (minimum 1 step) for a subset of samples ($p_{\text{sample}} = 0.7$):
\begin{equation}
    \medmath{s_k \rightarrow \langle\text{MASK}\rangle \text{ with prob. } p_{\text{sample}}, \quad k \sim \text{Uniform}(1, n).}
\end{equation}
This forces the student to infer the missing intermediate logic from the surrounding context, requiring genuine comprehension of the reasoning flow. The remaining 30\% undergo only shuffling, providing curriculum diversity. The combination ensures the model cannot rely on surface patterns alone.

The training prompt presents the corrupted sequence and asks the student to produce the complete, correctly ordered reasoning, as illustrated in Figure~\ref{fig:mask_shuffle_example}.

Different from self-supervised rewards in MR-RLVR~\cite{wang2025masked} which use masking/reordering for reward modeling during RL, our objective applies these techniques during the SFT phase to address capacity mismatch before any RL training begins. By enforcing the reconstruction of the logical chain, the student learns the structure of complex reasoning without being overwhelmed by the burden of verbatim memorization. This stage serves as a critical warmup, ensuring the student possesses the necessary structural priors for subsequent length optimization.

Let $\tilde{r}$ denote the \textbf{formatted input prompt} containing the reconstruction instruction and the corrupted reasoning chain (shuffled and optionally masked). The cross-entropy loss applies to reconstructing the complete, correctly-ordered teacher response $r_{\mathcal{T}}$:
\begin{equation}
    \mathcal{L}_{\text{Stage1}} = -\sum_{t} \log \pi_\theta(r_{\mathcal{T}}^{(t)} \mid \tilde{r}, r_{\mathcal{T}}^{(1:t-1)}).
\end{equation}
Here, $r_{\mathcal{T}}^{(t)}$ denotes the $t$-th token of the target teacher response, and the loss is computed autoregressively conditioned on the corrupted input $\tilde{r}$ and the preceding tokens $r_{\mathcal{T}}^{(1:t-1)}$.

In contrast to discriminative re-ordering models (e.g., Step-BERT~\cite{xia2025reasoning}) that only identify sequence indices, and standard next-token prediction that relies on local correlations, our generative reconstruction task forces the model to \textbf{internalize the semantic topology} of the reasoning chain, mastering the underlying logic rather than surface patterns.

\subsection{Stage 2: GRPO-Based Compression.}\label{sec:stage2}

\begin{figure}[t]
    \centering
    \begin{minipage}{\linewidth}
    \begin{tcolorbox}[colback=gray!10, colframe=black, size=small, width=0.92\linewidth, left=1mm, right=1mm, top=0.2mm, bottom=0.2mm, boxsep=0.2mm, title={\scriptsize\textbf{Training Sample Transformation Example}}, fontupper=\scriptsize]
    \textbf{Original Teacher CoT:} \\
    Step 1: Calculate total apples ($3+2=5$). \\
    Step 2: Subtract eaten apples ($5-1=4$). \\
    Step 3: Answer is 4.

    \vspace{0.08cm}
    \textbf{Transformed Input (Masking + Shuffling):} \\
    The following steps are disordered and incomplete. Please restore: \\
    $[$Step$]$ Subtract eaten apples ($5-1=4$) \hfill \textit{(Originally Step 2)} \\
    $[$Step$]$ $\langle$MASK$\rangle$ \hfill \textit{(Originally Step 1, masked)} \\
    $[$Step$]$ Answer is 4. \hfill \textit{(Originally Step 3)}

    \vspace{0.05cm}
    \textit{Goal: Restore original order and fill in the missing step.}
    \end{tcolorbox}
    \captionof{figure}{Illustration of the Structure-Aware Warmup data construction. We randomly mask one step (with $p=0.7$) and shuffle the sequence to force the student to learn logical dependencies.}
    \label{fig:mask_shuffle_example}
    \end{minipage}
\end{figure}

Stage 1 equips the student with structural understanding: the ability to recognize logical dependencies and reconstruct reasoning chains. However, this alone does not guarantee concise outputs. The student may still produce extended responses that faithfully replicate the teacher's lengthy style, since reconstruction training imposes no pressure toward brevity. To bridge this gap, Stage 2 introduces explicit optimization for compression while maintaining correctness.

Our goal is to find a student policy that maximizes:
\begin{equation}
    \max_{\theta} \mathbb{E}_{q \sim \mathcal{D}, r \sim \pi_\theta(\cdot|\tilde{r}_{\text{S2}})} \left[ \mathds{I}[\text{Correct}(r)] - \lambda \cdot |r| \right],
\end{equation}
where $r$ is the student's generated reasoning chain, $\text{Correct}(r)$ indicates whether the extracted answer matches the ground truth, $|r|$ measures the response length in tokens, $\mathds{I}[\cdot]$ is the indicator function, and $\lambda$ balances accuracy against brevity. Since this objective involves the non-differentiable indicator function, we employ reinforcement learning to optimize it.

To maintain training efficiency on compact models, we employ \textbf{Group Relative Policy Optimization (GRPO)}~\cite{shao2024deepseekmath}. Standard RLHF methods like PPO require a separate value model (critic), which doubles memory costs. GRPO eliminates this overhead by estimating baselines from group averages. During this stage, we construct masked inputs $\tilde{r}_{\text{S2}}$ without shuffling steps, by masking at least 2 steps for completion.

Designing an effective reward function is critical for GRPO-based optimization. A naive approach would linearly combine correctness and length penalties: $R = \alpha \cdot \mathds{I}[\text{Correct}] - \beta \cdot |r|$. However, this design has a fundamental flaw: the length penalty applies regardless of correctness, so an incorrect but very short output can score higher than a correct but moderately long one. This leads to \textbf{reward hacking}, where the model learns to produce minimal outputs that fail to solve problems.

To prevent this, we design a \textbf{hierarchical reward} that imposes a strict priority: correctness first, efficiency second. The key insight is that efficiency bonuses should only be meaningful when the answer is already correct. Our reward function takes the form:
\begin{equation}
    R(r_i) = R_{\text{base}}(r_i) + \mathds{I}[\text{Correct}(r_i)] \cdot R_{\text{eff}}(r_i).
    \label{eq:reward}
\end{equation}

\textit{Base Reward} $R_{\text{base}}$: This component establishes the correctness foundation. Incorrect answers receive a fixed penalty, and format violations (e.g., missing answer markers) receive an additional penalty. This ensures the model cannot bypass reasoning by producing malformed outputs. The base reward creates a clear separation: only outputs that pass this threshold can benefit from efficiency bonuses.

\textit{Efficiency Reward} $R_{\text{eff}}$: This component, gated by the correctness indicator $\mathds{I}[\text{Correct}]$, rewards compression only when the answer is correct. We define $R_{\text{eff}} = \gamma \cdot (1 - |r_i|/|r_{\text{baseline}}|)$, where $|r_{\text{baseline}}|$ is a reference length (e.g., teacher's output length). This relative formulation ensures the reward scales appropriately across problems of varying difficulty.

The multiplicative gating mechanism is crucial: an incorrect output receives zero efficiency reward regardless of its brevity, while a correct output is further rewarded for being brief. This design aligns the optimization landscape with our true objective and eliminates the reward hacking vulnerability. Specific reward coefficient values and their sensitivity analysis are provided in \textbf{Appendix~\ref{app:reward}}.

For each question $q$, the student generates $G$ candidate responses $\{r_1, \ldots, r_G\}$. The policy gradient update incorporates KL regularization to prevent distribution drift from the reference model $\pi_{\text{ref}}$ (initialized from Stage 1):
\begin{equation}
\begin{aligned}
    \nabla_{\theta} \mathcal{J} &= \mathbb{E} \left[ \sum_{i=1}^{G} (R(r_i) - \bar{R}) \nabla_{\theta} \log \pi_\theta(r_i \mid q) \right] \\
    &\quad - \beta_{\text{KL}} \nabla_{\theta} D_{\text{KL}}(\pi_\theta \| \pi_{\text{ref}}),
\end{aligned}
\end{equation}
where $\bar{R} = \frac{1}{G}\sum_j R(r_j)$ serves as the baseline and $\beta_{\text{KL}}$ controls the strength of KL regularization. This KL term prevents the model from deviating excessively from the structural knowledge acquired in Stage 1.

Stage 2 training continues until the accuracy-compression trade-off stabilizes. This stage also reveals which samples exceed the student's current capabilities, identifying failure cases for targeted enhancement in Stage 3.

\subsection{Stage 3: Teacher-Guided Internalization.}

Stage 2 successfully optimizes the accuracy-compression trade-off for most training samples. However, some difficult problems remain beyond the student's capabilities, even after GRPO optimization. For these failure cases, we design a teacher-guided internalization process that enables the student to absorb the teacher's reasoning logic and express it in its own compact style.

\paragraph{Failure Case Identification.} The \textbf{Stage 2} model generates responses for remaining training samples. We classify each as correct (retained) or incorrect (collected into $\mathcal{D}_{\text{hard}}$). This self-evaluation reveals which problems exceed the student's current capabilities. For \textbf{BRIDGE}, we specifically target these failure cases rather than the entire dataset.

\paragraph{Teacher-Scaffolded Generation.} For samples in $\mathcal{D}_{\text{hard}}$, we construct prompts that provide the teacher's complete solution as a scaffold. The student is asked to rewrite the reasoning concisely, as illustrated in Figure~\ref{fig:rewrite_prompt}. Appendix~\ref{app:prompts} (Figure~\ref{fig:prompt_stage3}) gives the exact Stage~3 prompt used in our implementation.

\begin{figure}[t]
    \centering
    \begin{tcolorbox}[colback=gray!10, colframe=black, size=small, width=0.92\linewidth, left=1mm, right=1mm, top=0.2mm, bottom=0.2mm, boxsep=0.2mm, title={\scriptsize\textbf{Internalization Prompt Template}}]
    \scriptsize
    \textbf{Input:} \\
    Question: [Problem Text] \\
    Teacher's Reference Solution: [Long verbose CoT...]

    \vspace{0.02cm}
    \textbf{Instruction:} \\
    Based on the reference, write a solution that is \textbf{concise} and \textbf{correct}. Do not copy word-for-word.

    \textit{Goal: Force the student to compress and internalize the teacher's logic.}
    \end{tcolorbox}
    \caption{Illustrative prompt template for the internalization step. The student sees the teacher's complete solution but must express the reasoning in its own concise style. See Appendix~\ref{app:prompts} (Figure~\ref{fig:prompt_stage3}) for the exact prompt used in our implementation.}
    \label{fig:rewrite_prompt}
\end{figure}

This is rewriting, not copying. The student sees the answer but must express the reasoning in its own words and style, selecting which steps to retain or merge. This process enables internalization: the student absorbs the teacher's logical structure while adapting it to its own capacity constraints.

\paragraph{Compression-Oriented Reward Design.} A key insight motivates our reward design: while students cannot generate verbose CoT from scratch for difficult problems, they retain sufficient capacity to comprehend and compress the logic when provided with structural guidance. In a preliminary experiment, \textbf{96.83\% (550/568)} of failure cases could produce at least one valid compression given teacher scaffolds. This observation suggests that reasoning redundancy in teacher outputs is largely linguistic overhead rather than logical need.

Based on this insight, we design a \textit{relative compression reward}. Following the same hierarchical structure as Stage 2 (Eq.~\ref{eq:reward}), the complete Stage 3 reward is:
\begin{equation}
    R(r_i) = R_{\text{base}}(r_i) + \mathds{I}[\text{Correct}(r_i)] \cdot R_{\text{comp}}(r_i),
    \label{eq:reward_s3}
\end{equation}
where $R_{\text{base}}$ ensures correctness (same as Stage 2). The compression bonus $R_{\text{comp}}$ explicitly encourages shorter-than-teacher outputs:
\begin{equation}
    R_{\text{comp}}(r_i) = \gamma \cdot \left(1 - \frac{|r_i|}{|r_{\mathcal{T}}|}\right),
\end{equation}
where $|r_{\mathcal{T}}|$ is the teacher's solution length. Unlike Stage 2, Stage 3 provides the complete teacher CoT as input, creating a risk of superficial copying. To prevent this, $R_{\text{comp}}$ assigns \textbf{negative} values when $|r_i| > |r_{\mathcal{T}}|$, penalizing outputs longer than the teacher while rewarding compression.

\paragraph{GRPO-Based Self-Discovery.} Rather than providing fixed rewritten solutions for memorization, we apply GRPO directly on teacher-scaffolded inputs, allowing the student to self-discover the optimal compression:
\begin{equation}
\resizebox{0.85\linewidth}{!}{$%
\begin{aligned}
    \nabla_\theta \mathcal{J}_{\text{S3}} &= \mathbb{E}_{\mathcal{D}_{\text{hard}}} \left[ \frac{1}{G} \sum_{i=1}^G \left( R(r_i) - \bar{R} \right) \nabla_\theta \log \pi_\theta(r_i \mid q, r_{\mathcal{T}}) \right] \\
    &\quad - \beta_{\text{KL}} \nabla_\theta D_{\text{KL}}(\pi_\theta \| \pi_{\text{S2}}),
\end{aligned}$%
}
\end{equation}
where the policy is conditioned on both the query $q$ and the teacher's scaffold $r_{\mathcal{T}}$. The reference model $\pi_{\text{S2}}$ is the Stage 2 model, preserving compression capabilities from training.

Crucially, we do not provide a ``correct'' short answer for supervision---the student must self-discover how to compress the verbose scaffold. While training uses teacher scaffolds, \textbf{inference requires only the query}: the scaffold is a learning tool that helps internalize concise reasoning into parameters.

The final \textbf{BRIDGE} model combines structural understanding from \textbf{Stage 1}, compression skills from \textbf{Stage 2}, and internalized reasoning with brevity incentives from \textbf{Stage 3}.

\subsection{Discussion.}

The \textbf{BRIDGE} curriculum is designed around a key principle: \textbf{internalization before compression}. Each stage addresses a specific aspect of the capacity mismatch problem:

\textbf{Stage 1} teaches the student to understand reasoning structure without requiring it to generate from scratch. By reconstructing shuffled and masked sequences, the student develops an internal model of logical dependencies, establishing the foundation for subsequent optimization.

\textbf{Stage 2} introduces compression pressure through GRPO on masked completion tasks. During this stage, we construct masked inputs $\tilde{r}_{\text{S2}}$ without shuffling steps, by masking at least 2 steps for completion, transitioning from passive reconstruction to active generation while the hierarchical reward ensures correctness before brevity.

\textbf{Stage 3} addresses the remaining hard cases through teacher-guided self-discovery. The observation that students can compress 96.83\% of teacher solutions reveals a fundamental asymmetry: generating verbose CoT from scratch is difficult, but compressing existing reasoning is within the student's capacity. BRIDGE exploits this asymmetry by providing teacher scaffolds while allowing the student to explore its own compression strategies.

This curriculum design embodies a shift from \textbf{memorization to internalization}. Rather than forcing the student to copy teacher outputs verbatim, BRIDGE progressively builds the student's ability to understand, compress, and ultimately internalize the underlying reasoning patterns.

\section{Experiments.}

We evaluate \textbf{BRIDGE} on mathematical reasoning benchmarks, demonstrating its effectiveness in achieving the accuracy-compression trade-off. We first describe the experimental setup, then present main results across multiple datasets and model architectures, followed by ablation studies validating our curriculum design.

\subsection{Experimental Setup.}

\paragraph{Datasets and Evaluation.} We use GSM8K~\cite{cobbe2021training} as the primary training and evaluation benchmark, containing 7,473 training and 1,319 test grade-school math problems. To assess generalization, we also evaluate on SVAMP~\cite{patel2021nlp} and MATH-500 in a zero-shot setting. We report Pass@1 accuracy (percentage of correctly answered problems using greedy decoding) and average output token count. Evaluation uses greedy decoding (temperature=0) with maximum 512 new tokens.

\paragraph{Implementation Details.} Our teacher is DeepSeek-R1-Distill-Qwen-14B, a 14-billion-parameter model fine-tuned for mathematical reasoning. The students are Qwen2.5-3B~\cite{qwen2025qwen25technicalreport} and Llama-3.2-3B-Base~\cite{grattafiori2024llama}, general-purpose models without task-specific instruction tuning. We focus on the 3B scale because larger models can directly clone teacher patterns, while 3B models exhibit the most severe capacity mismatch. All experiments are initialized from the Base model (not Instruct versions) to ensure fair comparison. \textbf{Stage 1} uses LoRA~\cite{hu2022lora} (rank 16, alpha 32) for efficient SFT; for \textbf{Stage 2} and \textbf{3}, we merge LoRA weights and perform full-parameter GRPO training. Training hyperparameters are in \textbf{Appendix~\ref{app:training}}.

\paragraph{Baselines.} We compare against: (1) \textbf{Base Model} and \textbf{Instruct Model}~\cite{qwen2025qwen25technicalreport,grattafiori2024llama}; (2) \textbf{Std-CoT KD}~\cite{magister2023teaching}: standard fine-tuning on teacher CoT; (3) \textbf{Short-CoT} and \textbf{Mix Distill.}~\cite{li2023mixed}: truncated or mixed-length CoT distillation; (4) \textbf{SuperRL}~\cite{liu2025superrl}: traditional SuperRL using SFT consolidation.

\subsection{Main Results.}

\vspace{-0.3em}
\begin{table*}[t]
\caption{Performance comparison on GSM8K, SVAMP, and MATH-500. We report accuracy (\%) and average output token length. Best results are \textbf{bold}, second best are \underline{underlined}.}
\label{tab:main}
\centering
\resizebox{0.95\textwidth}{!}{%
\scriptsize
\begin{tabular}{@{}l ccc ccc@{}}
\toprule
& \multicolumn{3}{c}{\textbf{Qwen 2.5-3B}} & \multicolumn{3}{c}{\textbf{Llama 3.2-3B}} \\
\cmidrule(lr){2-4} \cmidrule(lr){5-7}
\textbf{Method} & \textbf{GSM8K} & \textbf{SVAMP} & \textbf{MATH-500} & \textbf{GSM8K} & \textbf{SVAMP} & \textbf{MATH-500} \\
& Acc / Tok & Acc / Tok & Acc / Tok & Acc / Tok & Acc / Tok & Acc / Tok \\
\midrule
Base Model & 64.90 / 230 & 79.33 / 162 & \underline{36.40} / 371 & 10.99 / 227 & 20.67 / 192 & 5.33 / 155 \\
Instruct Model & 64.52 / 269 & 45.33 / 214 & 34.20 / 440 & \textbf{53.45} / 220 & \textbf{69.00} / 127 & \textbf{19.40} / 181 \\
\midrule
\multicolumn{7}{l}{\textit{Traditional KD methods}} \\
Std-CoT KD & 71.50 / 374 & \underline{82.00} / 161 & 26.60 / 367 & 23.12 / \underline{176} & 45.00 / 149 & 7.20 / 185 \\
Short-CoT & 39.42 / \textbf{165} & 42.10 / \underline{140} & 8.50 / \textbf{220} & 25.93 / 183 & 38.33 / 130 & 6.20 / 185 \\
Mix Distill. & 56.03 / 194 & 58.20 / 170 & 15.40 / \underline{280} & 24.26 / \textbf{135} & 24.67 / \underline{105} & 6.40 / \underline{149} \\
SuperRL & \underline{75.36} / 209 & 79.33 / 194 & 33.20 / 355 & \underline{50.49} / 221 & \underline{61.33} / 173 & 8.60 / 245 \\
\midrule
\textbf{BRIDGE (Ours)} & \textbf{76.19} / \underline{167} & \textbf{83.33} / \textbf{105} & \textbf{38.20} / 322 & 42.46 / 185 & 50.33 / \textbf{52} & \underline{16.00} / \textbf{107} \\
\bottomrule
\end{tabular}%
}
\end{table*}
\vspace{-0.3em}

Table~\ref{tab:main} presents results across two model architectures and three benchmarks. We organize our analysis by examining different baseline categories and their distinct limitations.

\paragraph{Comparison with Pretrained Models.} On Qwen 2.5-3B, the Base model achieves 64.90\% accuracy with 230 tokens. The Instruct model shows no improvement (64.52\%) despite additional instruction tuning, indicating that generic alignment does not address mathematical reasoning. On Llama 3.2-3B, the Instruct model performs significantly better (53.45\%) due to its extensive instruction-following pre-training, but this advantage comes from massive compute rather than efficient distillation.

\paragraph{Comparison with Distillation Methods.} Std-CoT KD improves accuracy to 71.50\% but produces excessively long outputs (374 tokens), demonstrating inefficient imitation of lengthy teacher patterns. Short-CoT achieves brevity (165 tokens) but severely degrades accuracy to 39.42\%, confirming that naive truncation destroys reasoning integrity. Mix Distill. attempts to balance both but achieves neither goal effectively (56.03\% / 194 tokens). These results highlight a fundamental tension in distillation: directly copying teacher outputs leads to verbosity, while forcing brevity destroys logic.

\paragraph{Comparison with RL Methods.} SuperRL achieves strong accuracy (75.36\%) but regresses on length (209 tokens). We attribute this to its reliance on SFT for consolidation, which overrides the compression incentives learned during RL. In contrast, BRIDGE avoids SFT in later stages, leveraging GRPO throughout Stage 3 to maintain the accuracy-brevity balance while internalizing reasoning.

\paragraph{BRIDGE Performance.} \textbf{BRIDGE} achieves the best accuracy-compression trade-off on Qwen (76.19\% / 167 tokens), outperforming all baselines in accuracy while using 27.4\% fewer tokens than the Base model. The hierarchical reward design prevents reward hacking (unlike Short-CoT), and the curriculum ensures stable optimization (avoiding the instability of direct RL). On Llama, \textbf{BRIDGE} attains competitive performance (42.46\% / 185 tokens on GSM8K) among self-trained methods, significantly outperforming Std-CoT KD (23.12\%) while maintaining compact outputs.

\paragraph{Zero-Shot Generalization.} Beyond in-domain performance, \textbf{BRIDGE} demonstrates strong transfer to unseen benchmarks. On SVAMP, \textbf{BRIDGE} achieves 83.33\% (vs. 79.33\% Base, +4.0\%) with only 105 tokens. On MATH-500, \textbf{BRIDGE} attains 38.20\% (vs. 36.40\% Base, +1.8\%). These gains emerge despite training exclusively on GSM8K, suggesting that the internalized reasoning patterns are generalizable problem-solving strategies rather than dataset-specific templates. The compression learned during training does not sacrifice generality; rather, it encourages transferable representations by forcing the model to distill essential reasoning.

\subsection{Ablation Study.}\label{sec:ablation}

\vspace{-0.3em}

\begin{table*}[t]
\centering
\begin{minipage}{0.48\textwidth}
\caption{Ablation study on GSM8K (Qwen 2.5-3B).}
\label{tab:ablation}
\centering
\resizebox{\textwidth}{!}{%
\begin{tabular}{lcc}
\toprule
\textbf{Variant} & \textbf{Accuracy (\%)} & \textbf{Tokens} \\
\midrule
\multicolumn{3}{l}{\textit{Stage Contributions}} \\
\textbf{Stage 1} Only (SFT) & 74.68 & 144 \\
\textbf{Stage 1 + 2} (w/o \textbf{Stage 3}) & 73.09 & 140 \\
\textbf{Full BRIDGE} & \textbf{76.19} & 167 \\
\midrule
\multicolumn{3}{l}{\textit{Component Analysis}} \\
w/o Masking (shuffle only) & 74.21 & 158 \\
w/o Shuffling (mask only) & 73.85 & 172 \\
\bottomrule
\end{tabular}%
}
\end{minipage}%
\hfill
\begin{minipage}{0.48\textwidth}
\caption{Error pattern comparison on GSM8K test set.}
\label{tab:errors}
\centering
\resizebox{\textwidth}{!}{%
\scriptsize
\begin{tabular}{lc}
\toprule
\textbf{Category} & \textbf{Count} \\
\midrule
Correct by both Base and \textbf{BRIDGE} & 739 \\
Fixed by \textbf{BRIDGE} (Base wrong) & 266 \\
Regressed by \textbf{BRIDGE} (Base correct) & 117 \\
Wrong by both & 197 \\
\midrule
Net improvement & +149 \\
\bottomrule
\end{tabular}%
}
\end{minipage}
\end{table*}
\vspace{-0.3em}

Table~\ref{tab:ablation} isolates the contribution of each component in \textbf{BRIDGE}. We analyze the results below.

\paragraph{Stage 1 Components.} Both masking and shuffling contribute to \textbf{Stage 1} effectiveness. Removing masking (shuffle only) reduces accuracy by 2.0 points, while removing shuffling (mask only) incurs a 2.3-point drop. The combination provides complementary training signals: shuffling forces global structure understanding, while masking requires local inference to complete missing steps.

\paragraph{Stage 3 Contribution.} The slight accuracy drop from \textbf{Stage 1} to \textbf{Stage 1+2} (74.68\% $\rightarrow$ 73.09\%) reflects the compression pressure introduced by the efficiency reward. \textbf{Stage 3} recovers this loss while maintaining brevity, adding 3.1 accuracy points over \textbf{Stage 2} alone (76.19\% vs 73.09\%). The teacher-guided internalization successfully transfers knowledge for samples beyond the student's initial capabilities.

\subsection{Error Analysis.}

\begin{figure*}[t]
\centering
\begin{minipage}{0.48\textwidth}
    \centering
    \includegraphics[width=\textwidth]{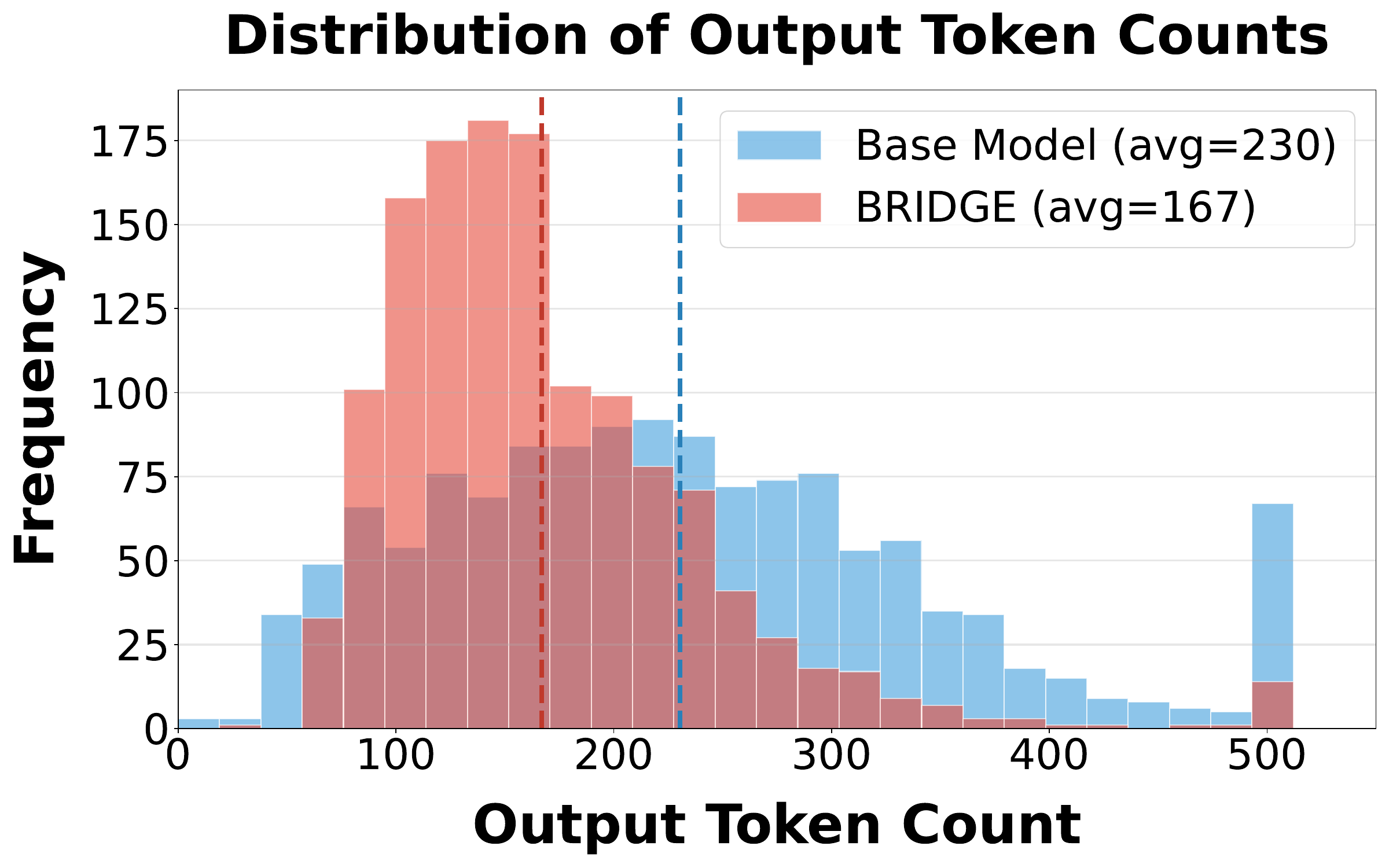}
    \caption{Output token distribution on GSM8K (Qwen 2.5-3B): Base model vs. \textbf{BRIDGE}.}
    \label{fig:tokens}
\end{minipage}%
\hfill
\begin{minipage}{0.48\textwidth}
    \centering
    \includegraphics[width=\textwidth]{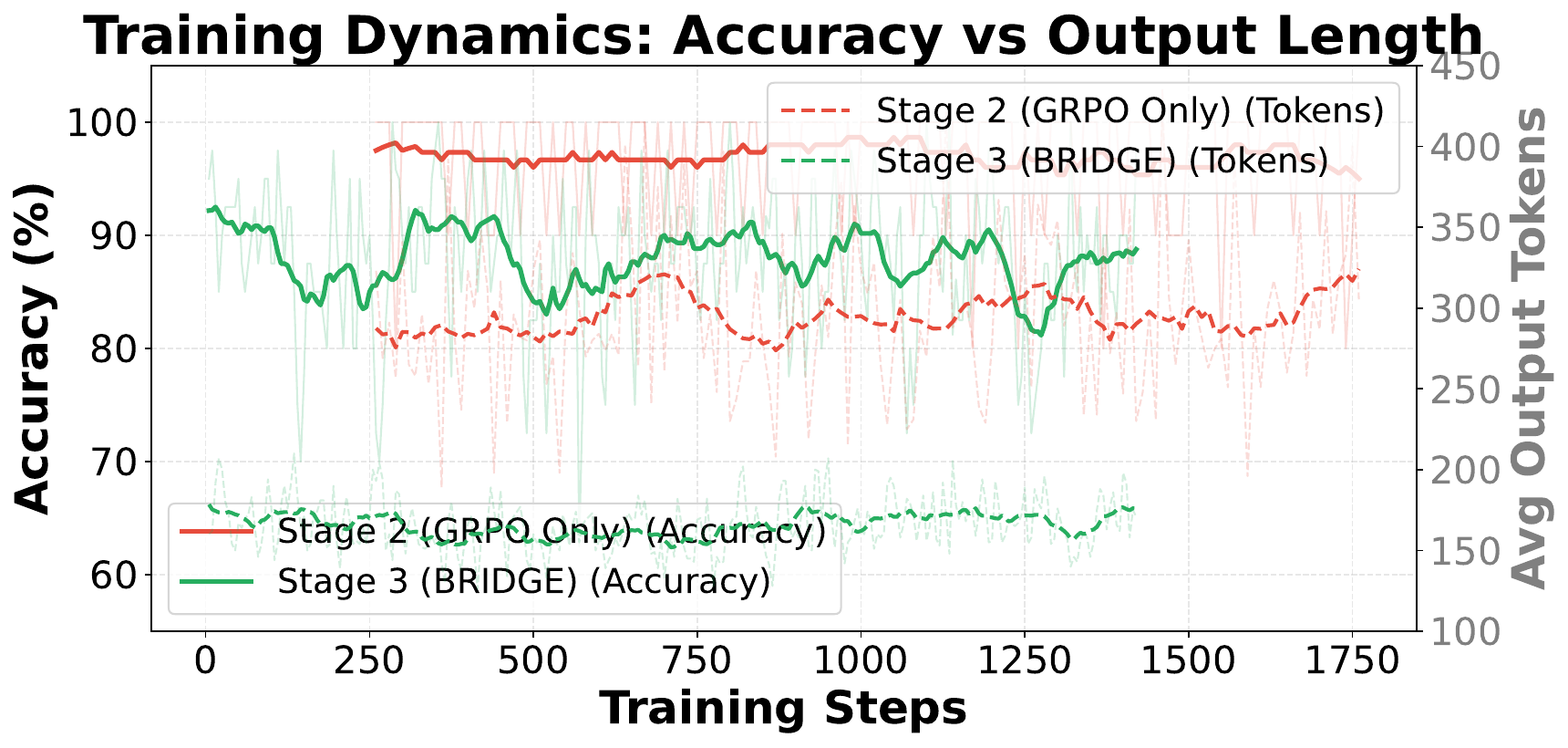}
    \caption{Training dynamics comparison. Solid lines: batch accuracy; dashed lines: output length. \textbf{BRIDGE} trains exclusively on 568 difficult samples (failure cases from Stage 2), explaining the lower batch accuracy but superior test performance.}
    \label{fig:trajectory}
\end{minipage}
\end{figure*}
\begin{figure*}[t!]
    \centering
    \begin{tcolorbox}[colback=gray!10, colframe=black, size=small, left=1mm, right=1mm, top=0.5mm, bottom=0.5mm, boxsep=0.5mm, title={\small\textbf{Case Study: Std-CoT KD vs. BRIDGE}}, width=0.92\linewidth]
    \small
    \textbf{Question:} Travis had 61 apps on his tablet. He deleted 9 apps he didn't use anymore and downloaded 18 more. How many apps are on his tablet now? \textbf{Gold Answer:} 70

    \vspace{0.05cm}
    \textbf{Std-CoT KD} (\textcolor{red}{Incorrect}, 512 tokens): ``Travis had 61 apps... He deleted 9 apps... downloaded 18 more...'' \textit{[repeats question 12 times, outputs \textbf{9}]}

    \vspace{0.05cm}
    \textbf{BRIDGE} (\textcolor{green!60!black}{Correct}, 80 tokens): Step 1: Travis started with 61 apps. Step 2: He deleted 9 apps: $61 - 9 = 52$. Step 3: He downloaded 18 more: $52 + 18 = 70$. Final Answer: \textbf{70}
    \end{tcolorbox}
    \caption{Std-CoT KD falls into repetition loops when overwhelmed, whereas \textbf{BRIDGE} generates concise, correct reasoning.}
    \label{fig:case_study}
\end{figure*}

Table~\ref{tab:errors} analyzes error patterns across the 1,319 GSM8K test samples. \textbf{BRIDGE} fixes 266 problems that the Base model answered incorrectly, while introducing 117 regressions. The net improvement of 149 samples accounts for the 11.29\% accuracy gain (76.19\% vs 64.90\%). Nearly all regressed samples stem from reasoning errors rather than formatting failures, with only a handful showing format issues such as repeated or missing steps, indicating room for further improvement in reasoning robustness.

\subsection{Efficiency Analysis.}

Figure~\ref{fig:tokens} compares output length distributions on GSM8K (Qwen 2.5-3B). The Base model exhibits a broad spread centered around 230 tokens, while \textbf{BRIDGE} produces a tighter distribution centered near 167 tokens. Notably, \textbf{BRIDGE} rarely generates outputs exceeding 300 tokens, whereas the Base model occasionally produces responses beyond 400 tokens. The compression is consistent across difficulties.
Figure~\ref{fig:case_study} illustrates a common failure mode: when trained with direct SFT on verbose teacher outputs, the model falls into degenerate repetition when overwhelmed. \textbf{BRIDGE}'s structure-aware training prevents this collapse by establishing reasoning competence before compression.

\subsection{Training Dynamics.}

Figure~\ref{fig:trajectory} compares training dynamics between \textbf{Stage 2} and the full \textbf{BRIDGE} pipeline. Although \textbf{BRIDGE} exhibits lower training batch accuracy, this reflects the increased difficulty of its training distribution---it exclusively processes samples that \textbf{Stage 2} failed to solve. Despite this harder curriculum, \textbf{BRIDGE} maintains stable output length with dramatically reduced token counts (167 vs 350). Critically, this translates to superior test performance: 76.19\% vs 73.09\%, showing hard sample targeting improves generalization.

\section{Related Work.}
\paragraph{Chain-of-Thought Distillation.}
Distilling reasoning capabilities from large teachers to small students is a rapidly evolving field. Early works like Std-CoT KD~\cite{magister2023teaching,ho2023large} demonstrated that fine-tuning on teacher rationales improves student performance. Recent work~\cite{hsieh2023distilling} further proposed multi-task learning to reduce data requirements, while Wang et al.~\cite{wang2023scott} introduced consistency constraints. However, these methods typically treat teacher outputs as immutable targets, leading to the capacity mismatch problem~\cite{li2025small}. Recent approaches attempt to address this by dynamically adjusting data. MCC-KD~\cite{chen2023mcc} and MiCoTA~\cite{ding2025micota} employ complex selection heuristics or teacher assistants, but they primarily target 7B+ students. In contrast, \textbf{BRIDGE} specifically targets the stricter constraints of 3B models by restructuring the learning process itself rather than just filtering data.

\paragraph{Reinforcement Learning for Reasoning.}
RLHF has been adapted for reasoning tasks to reduce dependency on gold labels. GRPO~\cite{shao2024deepseekmath} optimizes policy without a critic, reducing memory overhead. SuperRL~\cite{liu2025superrl} combines RL with SFT consolidation. However, our experiments show that standard SFT consolidation often overrides the brevity incentives learned during RL. \textbf{BRIDGE} addresses this by maintaining GRPO-based rewards throughout the internalization phase (\textbf{Stage 3}). While outcome-based RL is effective, it often lags behind process supervision~\cite{lightman2023let} in sample efficiency, and cold-start training on small models leads to unstable optimization~\cite{ouyang2022training}. Our \textbf{Stage 1} warmup provides crucial structural priors before RL begins, similar to pre-training before fine-tuning.

\paragraph{CoT Compression.}
Methods like TokenSkip~\cite{xia2025tokenskip} and entropy pruning~\cite{li2025compressing} compress CoT at inference time or via post-hoc pruning. An alternative line of work bypasses explicit reasoning entirely: implicit reasoning methods~\cite{deng2023implicit,li2025implicit,shen2025codi} compress reasoning into hidden states, but this sacrifices interpretability. Unlike both approaches, \textbf{BRIDGE} embeds compression directly into the model's generation capability during training while preserving explicit reasoning chains. Similarly, Mix Distill.~\cite{li2023mixed} mixes long and short data but relies on external heuristics. \textbf{BRIDGE} enables the student to \textit{self-discover} the optimal path via internalization. Moreover, post-hoc pruning methods~\cite{ma2023llm} often require architectural modifications or retraining. In contrast, \textbf{BRIDGE} achieves compression via parameter optimization, maintaining inference compatibility.

\section{Conclusion.}

We presented \textbf{BRIDGE}, a curriculum-based framework that addresses capacity mismatch in reasoning distillation by guiding student models from structure reconstruction (\textbf{Stage 1}) to compression via masked completion (\textbf{Stage 2}) and finally to teacher-guided internalization (\textbf{Stage 3}). The key insight is that students can compress teacher reasoning even when they cannot generate it from scratch---a capability we exploit through hierarchical rewards that prioritize correctness before efficiency. Experiments on GSM8K demonstrate that Qwen2.5-3B achieves 76.19\% accuracy with 167 tokens average output, compared to 64.90\% and 230 tokens for the original model, representing both accuracy gains and 27.4\% compression. Zero-shot transfer to SVAMP and MATH-500 confirms that the internalized reasoning patterns generalize beyond the training distribution. We discuss limitations and future directions in \textbf{Appendix~\ref{app:limitations}}.

\section*{Acknowledgments.}
This research was partially supported by National Natural Science Foundation of China (No.62502404), Hong Kong Research Grants Council (Research Impact Fund No.R1015-23, Collaborative Research Fund No.C1043-24GF, RGC Research Fellow Scheme No.RFS2627-1S03, General Research Fund No.~11218325, No.~11212926), Institute of Digital Medicine of City University of Hong Kong (No.9229503), Huawei (Huawei Innovation Research Program), Tencent (Tencent Rhino-Bird Focused Research Program, Tencent University Cooperation Project), Kuaishou (CCF-Kuaishou Large Model Explorer Fund No.~2025008, Kuaishou University Cooperation Project), Alibaba (CCF-Taobao \& Tmall Group TECH Kangaroo Fund 2026001), Didi (CCF-Didi Gaia Scholars Research Fund), and Bytedance.

\FloatBarrier

\bibliographystyle{siamplain}
\bibliography{references}

\clearpage
\appendix

\section{Training Details.}
\label{app:training}
\renewcommand{\thetable}{A.\arabic{table}}
\setcounter{table}{0}
\renewcommand{\thefigure}{A.\arabic{figure}}
\setcounter{figure}{0}

Table~\ref{tab:training_details} summarizes the hyperparameters used for each training stage.

\begin{table}[h]
    \centering
    \resizebox{0.92\linewidth}{!}{%
    \small
    \begin{tabular}{lccc}
        \toprule
        \textbf{Parameter} & \textbf{Stage 1} & \textbf{Stage 2} & \textbf{Stage 3} \\
        \midrule
        Training Samples & 2,000 & 2,000 & 568 \\
        Epochs / Steps & 5 epochs & ${\sim}$5,350 steps & 5 epochs \\
        Learning Rate & $2 \times 10^{-4}$ & $1 \times 10^{-5}$ & $1 \times 10^{-5}$ \\
        Batch Size & 8 & 8 & 8 \\
        LoRA Rank / Alpha & 16 / 32 & --- & --- \\
        GRPO Group Size $G$ & --- & 2 & 2 \\
        KL Coefficient $\beta_{\text{KL}}$ & --- & 0.1 & 0.1 \\
        PPO Update Epochs & --- & 4 & 4 \\
        \midrule
        GPU & \multicolumn{3}{c}{NVIDIA A100-80GB (single)} \\
        Precision & \multicolumn{3}{c}{bfloat16} \\
        \bottomrule
    \end{tabular}%
    }
    \caption{Training hyperparameters for \textbf{BRIDGE}.}
    \label{tab:training_details}
\end{table}

Stage 1 uses LoRA (applied to query and value projections) for efficient SFT. For Stage 2 and Stage 3, we merge the LoRA weights into the base model and perform full-parameter GRPO training to maximize optimization flexibility. We select the best checkpoint per stage based on validation performance.

\section{Complete Ablation Results.}
\label{app:ablation}
\renewcommand{\thetable}{B.\arabic{table}}
\setcounter{table}{0}
\renewcommand{\thefigure}{B.\arabic{figure}}
\setcounter{figure}{0}

Table~\ref{tab:complete_ablation} provides the full ablation results across all datasets and model architectures. These intermediate checkpoints (Stage 1 Only, Stage 2 Only) are not included in the main results table to maintain focus on comparing complete methods.

\begin{table}[h]
    \centering
    \resizebox{0.92\linewidth}{!}{%
    \small
    \begin{tabular}{l ccc}
        \toprule
        & \multicolumn{3}{c}{\textbf{Qwen 2.5-3B}} \\
        \cmidrule(lr){2-4}
        \textbf{Variant} & \textbf{GSM8K} & \textbf{SVAMP} & \textbf{MATH-500} \\
        & Acc / Tok & Acc / Tok & Acc / Tok \\
        \midrule
        Stage 1 Only & 74.68 / 144 & 81.20 / 130 & 31.40 / 490 \\
        Stage 1 + 2 & 73.09 / 140 & 80.50 / 120 & 32.80 / 413 \\
        Full \textbf{BRIDGE} & \textbf{76.19} / 167 & \textbf{83.33} / 105 & \textbf{38.20} / 322 \\
        \midrule
        & \multicolumn{3}{c}{\textbf{Llama 3.2-3B}} \\
        \cmidrule(lr){2-4}
        Stage 1 Only & 35.48 / 143 & 55.00 / 133 & 8.40 / 161 \\
        Stage 1 + 2 & 36.01 / 142 & 50.67 / 112 & 8.40 / 168 \\
        Full \textbf{BRIDGE} & 42.46 / 185 & 50.33 / 52 & \textbf{16.00} / 107 \\
        \bottomrule
    \end{tabular}%
    }
    \caption{Complete ablation results for \textbf{BRIDGE} stages across all benchmarks.}
    \label{tab:complete_ablation}
\end{table}

\section{Hyperparameter Sensitivity Analysis.}
\label{app:sensitivity}
\renewcommand{\thetable}{C.\arabic{table}}
\setcounter{table}{0}
\renewcommand{\thefigure}{C.\arabic{figure}}
\setcounter{figure}{0}

We conduct a sensitivity analysis on the group size $G$ in GRPO to verify the robustness of our hyperparameter choices. As shown in Table~\ref{tab:group_ablation}, $G=2$ achieves the best performance. While $G=4$ maintains similar accuracy with marginal degradation, $G=8$ shows significant performance drops in both accuracy and token efficiency, likely due to increased optimization complexity with larger sample groups. This validates our choice of $G=2$ as the optimal configuration.

\begin{table}[h]
    \centering
    \small
    \caption{Effect of GRPO group size $G$ on GSM8K performance.}
    \label{tab:group_ablation}
    \begin{tabular}{lccc}
        \toprule
        \textbf{Group Size} & \textbf{Accuracy (\%)} & \textbf{Tokens} & \textbf{Time} \\
        \midrule
        $G=2$ (Ours) & \textbf{76.19} & \textbf{167} & 1.0$\times$ \\
        $G=4$ & 75.89 & 166.9 & 2.0$\times$ \\
        $G=8$ & 73.84 & 211.8 & 4.0$\times$ \\
        \bottomrule
    \end{tabular}
\end{table}

Unlike standard GRPO implementations that require large group sizes for stable exploration~\cite{shao2024deepseekmath}, our curriculum framework provides strong structural priors in Stage 1, enabling efficient optimization with minimal sampling overhead.

\section{Reward Function Configuration.}
\label{app:reward}
\renewcommand{\thetable}{D.\arabic{table}}
\setcounter{table}{0}
\renewcommand{\thefigure}{D.\arabic{figure}}
\setcounter{figure}{0}

Table~\ref{tab:reward_config} details the hierarchical reward values used in Stage 2 and Stage 3 GRPO training. While both stages share the same hierarchical structure (correctness first, then efficiency), the efficiency bonus is computed differently.

\begin{table}[h]
    \centering
    \resizebox{0.92\linewidth}{!}{%
    \small
    \begin{tabular}{lcc}
        \toprule
        \textbf{Component} & \textbf{Stage 2} & \textbf{Stage 3} \\
        \midrule
        Incorrect Answer Penalty & $-2.0$ & $-2.0$ \\
        Format Error Penalty & $-1.0$ & $-1.0$ \\
        Correct Answer Base Reward & $+1.0$ & $+1.0$ \\
        \midrule
        \textit{Efficiency Bonus (if correct):} & & \\
        \quad Step Reduction Bonus (max) & $+0.9$ & --- \\
        \quad Token Efficiency Bonus (max) & $+0.2$ & --- \\
        \quad Compression Ratio Reward (max) & --- & $+0.8$ \\
        \midrule
        KL Coefficient ($\beta_{\text{KL}}$) & $0.1$ & $0.1$ \\
        Reference Model & Stage 1 & Stage 2 \\
        \bottomrule
    \end{tabular}%
    }
    \caption{Reward function configuration for GRPO training. Both stages share the base penalties but differ in efficiency bonus calculation.}
    \label{tab:reward_config}
\end{table}

Stage 2 rewards step reduction (fewer reasoning steps) and token efficiency relative to a baseline. Stage 3 focuses on compression ratio relative to the teacher's verbose solution, encouraging the student to distill verbose reasoning into concise formats.

\section{Complete Prompt Templates.}
\label{app:prompts}
\renewcommand{\thetable}{E.\arabic{table}}
\setcounter{table}{0}
\renewcommand{\thefigure}{E.\arabic{figure}}
\setcounter{figure}{0}

We provide all prompt templates used in our experiments. Figure~\ref{fig:prompt_teacher} through Figure~\ref{fig:prompt_inference} illustrate the complete prompts for each stage (the illustrative internalization template appears in the main text as Figure~\ref{fig:rewrite_prompt}).

\begin{figure}[H]
\centering
\begin{tcolorbox}[colback=gray!5, colframe=black!60, boxrule=0.5pt, width=0.9\linewidth]
Question: \{question\}\\
Answer:
\end{tcolorbox}
\caption{Teacher data generation prompt. DeepSeek-R1-Distill-Qwen-14B generates verbose chain-of-thought solutions. We filter for correct solutions, yielding 6,128 training samples.}
\label{fig:prompt_teacher}
\end{figure}

\begin{figure}[H]
\centering
\begin{tcolorbox}[colback=gray!5, colframe=black!60, boxrule=0.5pt, width=0.9\linewidth]
The following steps are out of order and incomplete, please complete and rearrange them.
\end{tcolorbox}
\caption{Stage 1 (Structure-Aware Reconstruction) prompt. Training uses masked and shuffled steps with this instruction.}
\label{fig:prompt_stage1}
\end{figure}

\begin{figure}[H]
\centering
\begin{tcolorbox}[colback=gray!5, colframe=black!60, boxrule=0.5pt, width=0.9\linewidth]
The following steps are incomplete, please complete them.
\end{tcolorbox}
\caption{Stage 2 (GRPO Compression) prompt. Training uses masked steps without shuffling.}
\label{fig:prompt_stage2}
\end{figure}

\begin{figure}[H]
\centering
\begin{tcolorbox}[colback=gray!5, colframe=black!60, boxrule=0.5pt, width=0.9\linewidth]
\small
You are given a math problem and its correct solution. Your task is to rewrite the solution step-by-step in a MUCH SHORTER way while keeping the same answer.\\[0.2em]

IMPORTANT: Your solution must be SHORTER than the teacher's solution. Combine steps, skip obvious calculations, and be concise.\\[0.2em]

Question: \{question\}\\[0.2em]

Teacher's Solution:\\
\{teacher\_cot\}\\[0.2em]

Now write a SHORTER solution. Use fewer steps. End with ``Final Answer: [number]''\\[0.2em]

Your Concise Solution:
\end{tcolorbox}
\caption{Stage 3 (Teacher-Guided Internalization) prompt. For failed samples from Stage 2, the student sees the teacher's complete solution and is prompted to rewrite it concisely.}
\label{fig:prompt_stage3}
\end{figure}

\begin{figure}[H]
\centering
\begin{tcolorbox}[colback=gray!5, colframe=black!60, boxrule=0.5pt, width=0.9\linewidth]
Question: \{question\}\\
Answer: Let me solve this step by step.
\end{tcolorbox}
\caption{Inference prompt used for all models (base, instruct, \textbf{BRIDGE}) for fair comparison.}
\label{fig:prompt_inference}
\end{figure}

\section{Error Analysis.}
\label{app:error}
\renewcommand{\thetable}{F.\arabic{table}}
\setcounter{table}{0}
\renewcommand{\thefigure}{F.\arabic{figure}}
\setcounter{figure}{0}

We analyze failure cases of \textbf{BRIDGE} on GSM8K to understand its limitations. Figure~\ref{fig:error_case} illustrates a typical failure mode.

\FloatBarrier

\begin{figure}[H]
\centering
\begin{tcolorbox}[colback=gray!5, colframe=black!60, boxrule=0.5pt, width=0.9\linewidth]
\textbf{Question:} A basket contains 25 oranges among which 1 is bad, 20\% are unripe, 2 are sour and the rest are good. How many good oranges are there?\\[0.3em]

\textbf{Gold Answer:} 17\\[0.3em]

\textbf{BRIDGE Output (145 tokens):}\\
Step 1: The basket contains 25 oranges.\\
Step 2: 20\% of 25 = 5 unripe oranges.\\
Step 3: There are 2 sour oranges.\\
Step 4: Good oranges = 25 - 5 - 2 = \textbf{18} $\leftarrow$ \textit{Missed the 1 bad orange}\\[0.3em]

\textbf{Predicted:} 18 \quad \textbf{Correct:} 17
\end{tcolorbox}
\caption{Example failure case showing condition omission error. The model skips the ``1 is bad'' condition during compression.}
\label{fig:error_case}
\end{figure}

\paragraph{Error Distribution.}
Among the 314 incorrect predictions (23.81\% of test set), we identify three main failure modes:
\begin{enumerate}
    \item \textbf{Condition Omission} (45\%): The model skips problem conditions during compression, as illustrated above.
    \item \textbf{Calculation Errors} (35\%): Arithmetic mistakes in intermediate steps.
    \item \textbf{Truncated Reasoning} (20\%): Generation stops before reaching the final answer.
\end{enumerate}

\section{Limitations and Future Work.}
\label{app:limitations}
\renewcommand{\thetable}{G.\arabic{table}}
\setcounter{table}{0}
\renewcommand{\thefigure}{G.\arabic{figure}}
\setcounter{figure}{0}

\paragraph{Limitations.} While we demonstrate zero-shot transfer to SVAMP and MATH-500, our training was conducted exclusively on the GSM8K dataset. Future work should explore training on broader, multi-domain reasoning corpora to further verify robustness. Additionally, the framework requires generating teacher CoT for training data, incurring computational costs that scale with dataset size. For even smaller models (e.g., 1.5B parameters), the Stage 1 masking difficulty may require adjustment---reducing the masking rate to accommodate more limited reasoning capacity---suggesting an interesting direction for capacity-aware curriculum design.

\paragraph{Future Work.} Several directions merit exploration. Extending \textbf{BRIDGE} to other reasoning domains---commonsense, logical, or scientific reasoning---would assess generalization. Investigating intermediate-sized teachers may reveal optimal teacher-student pairings. Combining our approach with inference-time compression methods could yield further efficiency gains. Finally, exploring iterative self-evolution, where the student eventually becomes its own teacher and continuously refines its internalized reasoning paths without external supervision, presents a promising direction toward fully autonomous reasoning improvement.

\end{document}